\documentclass[10pt,twocolumn,letterpaper]{article}

\usepackage{iccv}
\usepackage{times}
\usepackage{epsfig}
\usepackage{graphicx}
\usepackage{amsmath}
\usepackage{amssymb}
\usepackage[dvipsnames]{xcolor}
\usepackage{multirow}

\usepackage{colortbl}
\include{xspace}

\usepackage{pifont}
\newcommand{\cmark}{\textcolor{ForestGreen}{\ding{51}}}%
\newcommand{\xmark}{\textcolor{Red}{\ding{55}}}%

\definecolor{todo_color}{rgb}{.8,.3,.1}

\newcommand{\programmer}{program generator\xspace}
\newcommand{\interpreter}{execution engine\xspace}
\newcommand{\Programmer}{Program generator\xspace}
\newcommand{\Interpreter}{Execution engine\xspace}
\newcommand{\app}{\raise.17ex\hbox{$\scriptstyle\sim$}}
\newcommand{\Scene}{{\small\texttt{Scene}}\xspace}

\usepackage[pagebackref=true,breaklinks=true,letterpaper=true,colorlinks,bookmarks=false]{hyperref}

\iccvfinalcopy 

\def\iccvPaperID{****}

\ificcvfinal\fi

\begin{document}

\title{Inferring and Executing Programs for Visual Reasoning}

\author{
  Justin Johnson\textsuperscript{1} \hspace{2.25pc}
  Bharath Hariharan\textsuperscript{2} \hspace{2.25pc}
  Laurens van der Maaten\textsuperscript{2} \\
  Judy Hoffman\textsuperscript{1} \hspace{1pc}
  Li Fei-Fei\textsuperscript{1} \hspace{1pc}
  C. Lawrence Zitnick\textsuperscript{2} \hspace{1pc}
  Ross Girshick\textsuperscript{2}
  \\*[8pt]
  \begin{minipage}{0.4\textwidth}
    \centering
    \textsuperscript{1}Stanford University
  \end{minipage}
  \begin{minipage}{0.4\textwidth}
    \centering
    \textsuperscript{2}Facebook AI Research
  \end{minipage}
}

\maketitle

\begin{abstract}
  Existing methods for visual reasoning attempt to directly map inputs to outputs using black-box architectures without explicitly modeling the underlying reasoning processes. 
As a result, these black-box models often learn to exploit biases in the data rather than learning to perform visual reasoning. 
Inspired by module networks, this paper proposes a model for visual reasoning that consists of a \emph{\programmer} that constructs an explicit representation of the reasoning process to be performed, and an \emph{\interpreter} that executes the resulting program to produce an answer. 
Both the \programmer and the \interpreter are implemented by neural networks, and are trained using a combination of backpropagation and REINFORCE. 
Using the CLEVR benchmark for visual reasoning, we show that our model significantly outperforms strong baselines and generalizes better in a variety of settings.


\end{abstract}
\vspace{-3mm}

\section{Introduction}
 
\begin{figure}
  \centering
 \resizebox{0.96\linewidth}{!}{
    \begin{tabular}{cc}
  \includegraphics[width=.52\linewidth]{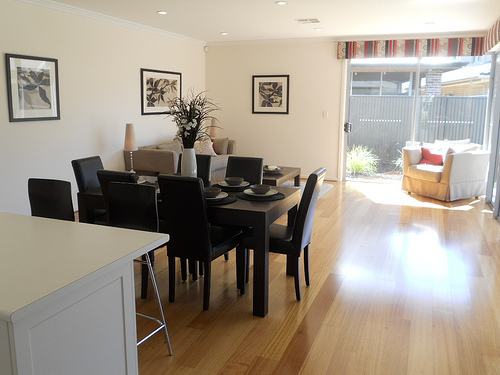}&
  \includegraphics[width=.52\linewidth]{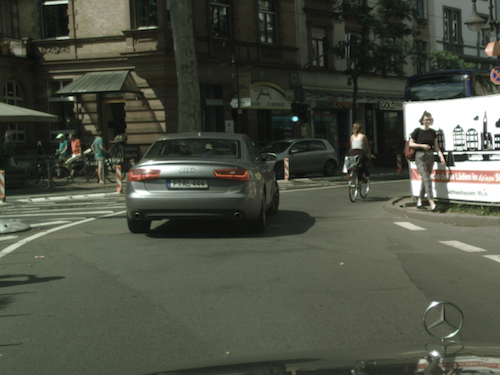}\\
  \begin{minipage}[b]{.52\linewidth}
	   \small{\emph{How many chairs are at the table?}}
  \end{minipage} & 
  \begin{minipage}[b]{.52\linewidth}
	   \small{\emph{Is there a pedestrian in my lane?}}
  \end{minipage} \\
  ~\\
  \includegraphics[width=.52\linewidth]{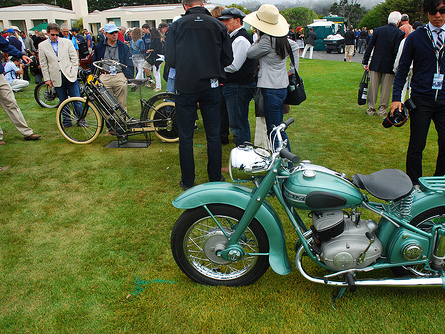}&
  \includegraphics[width=.52\linewidth]{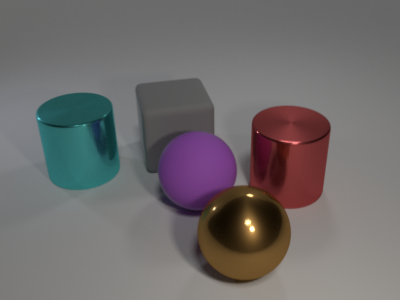}\\
  \begin{minipage}[b]{.52\linewidth}
	  \small{\emph{Is the person with the blue hat touching the bike in the back?}}
  \end{minipage}&
  \begin{minipage}[b]{.52\linewidth}
	  \small{\emph{Is there a matte cube that has the same size as the red metal object?}}
  \end{minipage} \\
  ~\\
  \end{tabular}
}
  \vspace{-1mm}
  \caption{
  Compositional reasoning is a critical component needed for understanding the complex visual scenes encountered in applications such as robotic navigation, autonomous driving, and surveillance. Current models fail to do such reasoning~\cite{johnson2017clevr}.
  }
  \vspace{-2mm}
  \label{fig:teaser}
\end{figure}

In many applications, computer-vision systems need to answer sophisticated queries by \emph{reasoning} about the visual world (Figure~\ref{fig:teaser}).
To deal with novel object interactions or object-attribute combinations, visual reasoning needs to be \emph{compositional}: without ever having seen a ``person touching a bike'', the model should be able to understand the phrase by putting together its understanding of ``person'', ``bike" and ``touching''.
Such compositional reasoning is a hallmark of human intelligence, and allows people to solve a plethora of problems using a limited set of basic skills~\cite{lake2016building}.

In contrast, modern approaches to visual recognition learn a mapping directly
from inputs to outputs; they do not explicitly formulate and execute
compositional plans.
Direct input-output mapping works well for classifying images~\cite{krizhevsky2012imagenet} and detecting objects~\cite{girshick2015fast} for a small, fixed set of categories.
However, it fails to outperform strong baselines on tasks that require the model to understand an exponentially large space of objects, attributes, actions, and interactions, such as visual question answering (VQA)~\cite{antol2015vqa,zhu2016v7w}.
Instead, models that learn direct input-output mappings tend to learn dataset biases but not reasoning~\cite{devlin2015exploring,jabri2016revisiting,johnson2017clevr}.

In this paper, we argue that to successfully perform complex reasoning tasks, it might be necessary to explicitly incorporate compositional reasoning in the model structure.
Specifically, we investigate a new model for visual question answering that consists of two parts: a \emph{\programmer} and an \emph{\interpreter}.
The \emph{\programmer} reads the question and produces a plan or \emph{program} for answering the question by composing functions from a function dictionary.
The \emph{\interpreter} implements each function using a small neural module, and executes the resulting module network on the image to produce an answer.
Both the \programmer and the modules in the \interpreter are neural networks with generic architectures; they can be trained separately when ground-truth programs are available, or jointly in an end-to-end fashion.

Our model builds on prior work on neural module networks that incorporate compositional reasoning~\cite{andreas2016learning,andreas2016neural}. Prior module networks do not generalize well to new problems, because they rely on a hand-tuned \programmer based on syntactic parsing, and on hand-engineered modules. By contrast, our model does not rely on such heuristics: we only define the function vocabulary and the ``universal'' module architecture by hand, learning everything else. 

We evaluate our model on the recently released CLEVR dataset~\cite{johnson2017clevr}, which has proven to be challenging for state-of-the-art VQA models. 
The CLEVR dataset contains ground-truth programs that describe the compositional reasoning required to answer the given questions.
We find that with only a small amount of reasoning supervision ($9000$ ground truth programs which is $2\%$ of those available), our model outperforms state-of-the-art non-compositional VQA models by $\app20$ percentage points on CLEVR.
We also show that our model's compositional nature allows it to generalize to novel questions by composing modules in ways that are not seen during training.

Though our model works well on the algorithmically generated questions in CLEVR, the true test is whether it can answer questions asked by humans in the wild.
We collect a new dataset of human-posed free-form natural language questions about CLEVR images.
Many of these questions have out-of-vocabulary words and require reasoning skills that are absent from our model's repertoire.
Nevertheless, when finetuned on this dataset without additional program supervision, our model learns to compose its modules in novel but intuitive ways to best answer new types of questions.
The result is an interpretable mapping of free-form natural language to programs, and a $\app9$ point improvement in accuracy over the best competing models.

\section{Related Work}
Our work is related to to prior research on visual question answering, reasoning-augmented models, semantic parsers, and (neural) program-induction methods.

\textbf{Visual question answering} (VQA) is a popular proxy task for gauging the quality of visual reasoning systems \cite{kafle16vqareview,wu16vqareview}. Like the CLEVR dataset, benchmark datasets for VQA typically comprise a set of questions on images with associated answers \cite{antol2015vqa,malinowski14multi,tapaswi2016movieqa,krishna2017visual,zhu2016v7w}; both questions and answers are generally posed in natural language. Many systems for VQA employ a very similar architecture \cite{antol2015vqa,fukui16mcb,gao2016compact,lu16coattention,malinowski15neurons,mallya2016vqa,xiong2016dynamic}: they combine an RNN-based embedding of the question with a convolutional network-based embedding of an image in a classification model over possible answers. Recent work has questioned whether such systems are capable of developing visual reasoning capabilities: (1) very simple baseline models were found to perform competitively on VQA benchmarks by exploiting biases in the data \cite{jabri2016revisiting,zhang16yinyang,goyal2017making} and (2) experiments on CLEVR, which was designed to control such biases, revealed that current systems do not learn to reason about spatial relationships or to learn disentangled representations \cite{johnson2017clevr}. 

Our model aims to address these problems by explicitly constructing an intermediate program that defines the reasoning process required to answer the question. We show that our model succeeds on several kinds of reasoning where other VQA models fail. 

\textbf{Reasoning-augmented models} add components to neural network models to facilitate the development of reasoning processes in such models. For example, models such as neural Turing machines \cite{graves2014neural,graves2016hybrid}, memory networks \cite{weston2015memory,sukhbaatar2015endtoend}, and stack-augmented recurrent networks \cite{joulin2015inferring} add explicit memory components to neural networks to facilitate learning of reasoning processes that involve long-term memory. While long-term memory is likely to be a crucial component of intelligence, it is not a prerequisite for reasoning, especially the kind of reasoning that is required for answering questions about images.\footnote{Memory is likely indispensable in more complex settings such as visual dialogues or SHRDLU \cite{das2016dialogue,winograd72}.} Therefore, we do not consider memory-augmented models in this study.

Module networks are an example of reasoning-augmented models that use a syntactic parse of a question to determine the architecture of the network \cite{andreas2016learning,andreas2016neural,hu2016modeling}. The final network is composed of trained neural modules that execute the ``program'' produced by the parser. The main difference between our models and existing module networks is that we replace hand-designed off-the-shelf syntactic parsers \cite{klein03parser}, which perform very poorly on complex questions such as those in CLEVR \cite{johnson2017clevr}, by a learnt \programmer that can adapt to the task at hand.

\textbf{Semantic parsers} attempt to map natural language sentences to logical forms. 
Often, the goal is to answer natural language questions using a knowledge base~\cite{liang2011learning}.
Recent approaches to semantic parsing involve a learnt \emph{programmer}~\cite{liang2016neural}.
However, the semantics of the program and the execution engine are fixed and known a priori, while we learn both the \programmer and the \interpreter. 

\textbf{Program-induction methods} learn programs from input-output pairs by fitting the parameters of a neural network to predict the output that corresponds to a particular input value. Such models can take the form of a feedforward scoring function over operators in a domain-specific language that can be used to guide program search \cite{balog2017deepcoder}, or of a recurrent network that decodes a vectorial program representation into the actual program \cite{kaiser2016neural,kurach2016neural,neelakantan2016neural,zaremba2016learning,zaremba2014learning,zaremba2015reinforcement}. The recurrent networks may incorporate compositional structure that allows them to learn new programs by combining previously learned sub-programs \cite{reed2016neural}.

Our approach differs from prior work on program induction in (1) the type of input-output pairs that are used and (2) the way the domain-specific language is implemented. Prior work on neural program interpreters considers simple algorithms such as sorting of a list of integers; by contrast, we consider inputs that comprise an image and an associated question (in natural language). Program induction approaches also assume knowledge of the low-level operators such as arithmetic operations. In contrast, we use a learnt \interpreter and assume minimal prior knowledge.

\section{Method}
We develop a learnable compositional model for visual question answering. 
Our model takes as input an image $x$ and a visual question $q$ about the image.
The model selects an answer $a\in\mathcal{A}$ to the question from a fixed set $\mathcal{A}$
of possible answers. Internally, the model predicts a program $z$ representing the reasoning
steps required to answer the question. The model then executes the predicted program on the
image, producing a distribution over answers.

To this end, we organize our system into two components: a \emph{\programmer},
$z=\pi(q)$, which predicts programs from questions, and an 
\emph{\interpreter}, $a=\phi(x, z)$, which executes a program $z$ on an image $x$
to predict an answer $a$.
Both the \programmer and the \interpreter are neural networks that are
learned from data. In contrast to prior work~\cite{andreas2016learning,andreas2016neural},
we do not manually design heuristics for generating or executing the programs. 

We present learning procedures both for settings where (some) ground-truth programs are available during training, and for settings without ground-truth programs. 
In practice, our models need \emph{some} program supervision during training, but we find that the \programmer requires very few of such programs in order to learn to generalize (see Figure~\ref{fig:accvsdata}).

\subsection{Programs}
 Like all programming languages, our programs are defined by \emph{syntax} giving rules for building valid programs,
 and \emph{semantics} defining the behavior of valid programs.
 We focus on learning semantics for a fixed syntax. Concretely, we fix the syntax by pre-specifying a set $\mathcal{F}$ of functions $f$,
each of which has a fixed arity $n_f \in \{1,2\}$. Because we are 
interested in visual question answering, we include in the vocabulary a 
special constant \Scene, which represents the visual features of
the image. 
We represent valid programs $z$ as \emph{syntax trees} in which 
each node contains a function $f \in \mathcal{F}$, and in which each node has as many children as the arity of the function $f$.

\begin{figure}
  \centering
  \includegraphics[width=\linewidth]{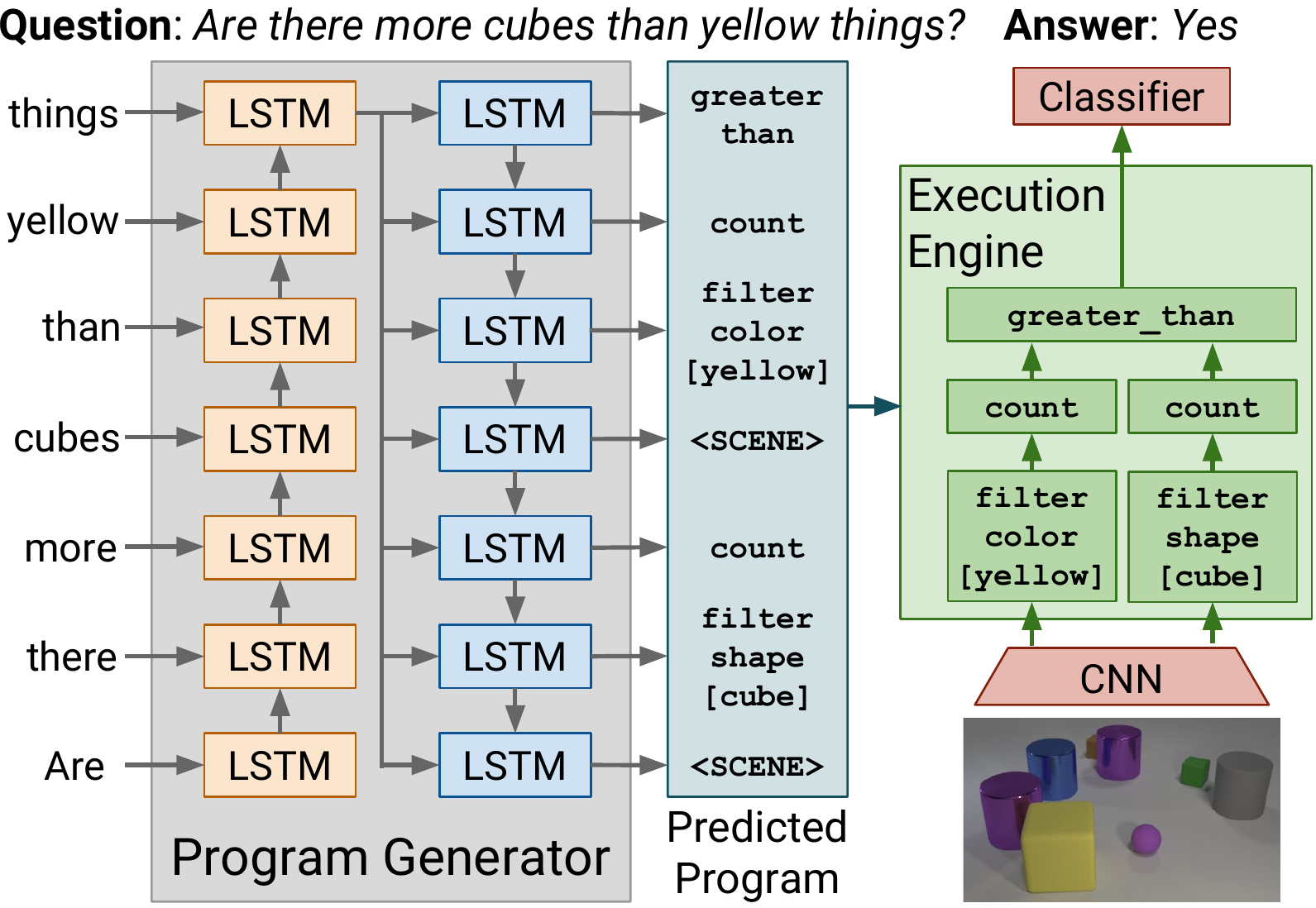}\vspace{.05mm}
  \caption{
    System overview. The \textbf{\programmer} is a sequence-to-sequence model
    which inputs the question as a sequence of words and outputs a program as
    a sequence of functions, where the sequence is interpreted as a
    prefix traversal of the program's abstract syntax tree.
    The \textbf{\interpreter} executes the program on the image by assembling
    a neural module network~\cite{andreas2016neural} mirroring the structure
    of the predicted program.
  }
  \label{fig:system}
\end{figure}

\subsection{\Programmer}
The \programmer $z=\pi(q)$ predicts programs $z$ from natural-language questions
$q$ that are represented as a sequence of words.
We use a prefix traversal to serialize the syntax tree, which is a non-sequential discrete structure, into a sequence of functions. This allows us to implement the \programmer
using a standard LSTM sequence-to-sequence model; see~\cite{sutskever2014sequence} for details.

When decoding at test time, we simply take the argmax function at each time step.
The resulting sequence of functions is converted
to a syntax tree; this is straightforward since the arity of each function is known.
Some generated sequences do not correspond to prefix traversals of a tree. If the
sequence is too short (some functions do not have enough children) then we
pad the sequence with \Scene constants. If the sequence is too long (some functions
have no parents) then unused functions are discarded.

\begin{table*}
  \small
  \centering
  \renewcommand{\arraystretch}{1.2}
\renewcommand{\tabcolsep}{1.2mm}
\resizebox{0.95\linewidth}{!}{
  \begin{tabular}{r|cc|ccc|cccc|cccc|c}
    & & & \multicolumn{3}{c|}{Compare Integer} & \multicolumn{4}{c|}{Query} & \multicolumn{4}{c|}{Compare} \\
    Method & Exist & Count & Equal & Less & More & Size & Color & Mat. & Shape & Size & Color & Mat. & Shape & Overall \\
    \hline\hline
    Q-type mode & 50.2 & 34.6 & 51.4 & 51.6 & 50.5 & 50.1 & 13.4 & 50.8 & 33.5 & 50.3 & 52.5 & 50.2 & 51.8 & 42.1 \\ 
    LSTM & 61.8 & 42.5 & 63.0 & 73.2 & 71.7 & 49.9 & 12.2 & 50.8 & 33.2 & 50.5 & 52.5 & 49.7 & 51.8 & 47.0 \\
    CNN+LSTM & 68.2 & 47.8 & 60.8 & 74.3 & 72.5 & 62.5 & 22.4 & 59.9 & 50.9 & 56.5 & 53.0 & 53.8 & 55.5 & 54.3 \\ 
    CNN+LSTM+SA~\cite{yang16stackedattention} & 68.4 & 57.5 & 56.8 & 74.9 & 68.2 & 90.1 & 83.3 & 89.8 & 87.6 & 52.1 & 55.5 & 49.7 & 50.9 & 69.8 \\ 
    CNN+LSTM+SA+MLP & 77.9 & 59.7 & 60.3 & 83.7 & 76.7 & 85.4 & 73.1 & 84.5 & 80.7 & 72.3 & 71.2 & 70.1 & 69.7 & 73.2 \\ 
    \hline
    Human$^\dagger$~\cite{johnson2017clevr} & 96.6 & 86.7 & 79.0 & 87.0 & 91.0 & 97.0 & 95.0 & 94.0 & 94.0 & 94.0 & 98.0 & 96.0 & 96.0 & 92.6 \\
    \hline
    Ours-strong (700K prog.) & \bf{97.1} & \bf{92.7} & \bf{98.0} & \bf{99.0} & \bf{98.9} & \bf{98.8} & \bf{98.4} & \bf{98.1} & \bf{97.3} & \bf{99.8} & \bf{98.5} & \bf{98.9} & \bf{98.4} & \bf{96.9} \\ 
    Ours-semi (18K prog.) & 95.3 & 90.1 & 93.9 & 97.1 & 97.6 & 98.1 & 97.1 & 97.7 & 96.6 & 99.0 & 97.6 & 98.0 & 97.3 & 95.4 \\ 
    Ours-semi (9K prog.) & 89.7 & 79.7 & 85.2 & 76.1 & 77.9 & 94.8 & 93.3 & 93.1 & 89.2 & 97.8 & 94.5 & 96.6 & 95.1 & 88.6 \\ 
    
  \end{tabular}}
  \vspace{1mm}
  \caption{
    Question answering accuracy (higher is better) on the CLEVR dataset for baseline models, humans, and three variants of our model. The strongly supervised variant of our model uses all 700K ground-truth programs for training, whereas the semi-supervised variants use 9K and 18K ground-truth programs, respectively. $^\dagger$Human performance is measured on a 5.5K subset of CLEVR questions.
  }
  \vspace{-2mm}
  \label{tab:fullclevr}
\end{table*}

\subsection{\Interpreter}
Given a predicted program $z$ and and an input image $x$, the \interpreter executes the program
on the image, $a=\phi(x, z)$, to predict an answer $a$. The \interpreter is implemented using a neural module network~\cite{andreas2016neural}: the program $z$ is used to assemble a question-specific neural network
that is composed from a set of modules. For each function $f \in \mathcal{F}$, the \interpreter maintains a
neural network module $m_f$. Given a program $z$, the \interpreter creates a neural
network $m(z)$ by mapping each function $f$ to its corresponding module $m_f$ in the order defined by the program: the outputs of the ``child modules'' are used as input into their corresponding ``parent module''.

Our modules use a generic architecture, in contrast to~\cite{andreas2016neural}. A module of arity $n$ receives $n$ features maps of shape
$C{\times}H{\times}W$ and produces a feature map of shape $C{\times}H{\times}W$. Each unary module is a
standard residual block~\cite{he2016deep} with two $3\times3$ convolutional layers.
Binary modules concatenate their inputs along the channel dimension, project from $2C$ to $C$ channels
using a $1\times1$ convolution, and feed the result to a residual block. 
The \Scene module takes visual features as input ($\mathit{conv}4$ features from
ResNet-101~\cite{he2016deep} pretrained on ImageNet~\cite{russakovsky2015imagenet}) and passes
these features through four convolutional layers to output a $C{\times}H{\times}W$ feature map.

Using the same architecture for all modules ensures that every valid program $z$ corresponds to a
valid neural network which inputs the visual features of the image and outputs a feature map of
shape $C{\times}H{\times}W$. This final feature map is flattened and passed into a multilayer perceptron classifier that outputs a distribution over possible answers.

\subsection{Training}
\label{sec:training}
Given a VQA dataset containing $(x, q, z, a)$ tuples with ground truth programs $z$,
we can train both the \programmer and \interpreter in a supervised manner.
Specifically, we can (1) use pairs $(q, z)$ of questions and corresponding programs to train the
\programmer, which amounts to training a standard sequence-to-sequence model; and (2)
use triplets $(x,z,a)$ of the image, program, and answer to train the \interpreter,
using backpropagation to compute the required gradients (as in~\cite{andreas2016neural}).

Annotating ground-truth programs for free-form natural language questions is expensive,
so in practice we may have few or no ground-truth programs. To address this problem, we
opt to train the \programmer and \interpreter jointly on $(x, q, a)$ triples \emph{without
ground-truth programs}. However, we cannot backpropagate through the argmax operations in the
\programmer. Instead we replace the argmaxes with sampling and use REINFORCE~\cite{williams1992} to estimate gradients on the outputs
of the \programmer; the reward for each of its outputs is the negative zero-one loss of
the \interpreter, with a moving-average baseline.

In practice, joint training using REINFORCE is difficult: the \programmer needs to produce the right program 
without understanding what the functions mean, and the \interpreter has to produce the right 
answer from programs that may not accurately implement the question asked. We propose a more practical
\emph{semi-supervised learning} approach. We first use a \emph{small} set of ground-truth programs to train the \programmer,
then fix the \programmer and train the \interpreter using predicted programs on a large dataset of $(x, q, a)$ triples.
Finally, we use REINFORCE to jointly finetune the \programmer and \interpreter. Crucially, ground-truth programs are \emph{only}
used to train the initial \programmer.


\section{Experiments}

\newcommand{\subwidth}{0.19\linewidth}
\begin{figure*}
  \small\centering
  \begin{minipage}{\subwidth}\centering \textbf{Q:} \textit{What shape is the}\dots \\* ~\\~\\*\end{minipage}
  \begin{minipage}{\subwidth}
    \small\centering
    \dots\underline{\smash{\textcolor{Plum}{purple}}}\textit{ thing?} \\~\\ 
    \textbf{A:} \textit{cube}\\
  \end{minipage}
  \begin{minipage}{\subwidth}
    \small\centering
    \dots\underline{\textcolor{RoyalBlue}{blue}}\textit{ thing?} \\~\\
    \textbf{A:} \textit{sphere}
  \end{minipage}
  \begin{minipage}{\subwidth}
    \small\centering
    \dots\underline{\textcolor{red}{red}}\textit{ thing }\underline{\smash{right}}\textit{ of \\* the \textcolor{RoyalBlue}{blue} thing?} \\*
    \textbf{A:} \textit{sphere}
  \end{minipage}
  \begin{minipage}{\subwidth}
    \small\centering
    \dots\underline{\textcolor{red}{red}}\textit{ thing }\underline{\smash{left}}\textit{ of \\* the \textcolor{RoyalBlue}{blue} thing?} \\*
    \textbf{A:} \textit{cube}
  \end{minipage}\\*
  \includegraphics[width=\subwidth]{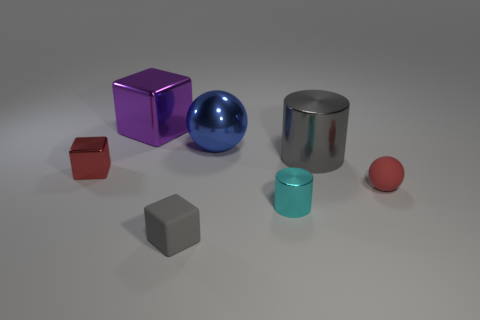} 
  \includegraphics[width=\subwidth]{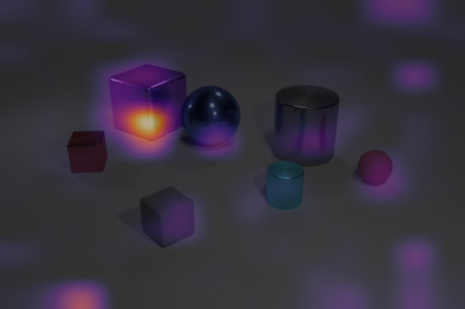} 
  \includegraphics[width=\subwidth]{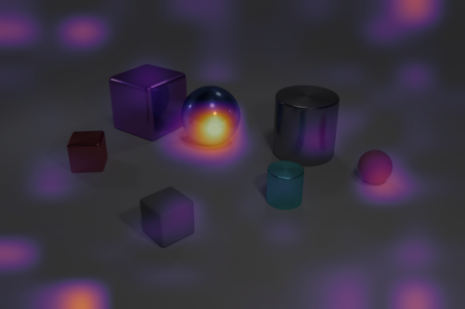} 
  \includegraphics[width=\subwidth]{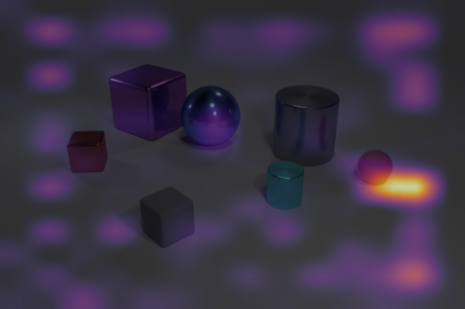} 
  \includegraphics[width=\subwidth]{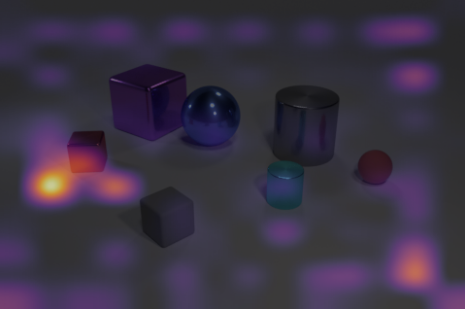} \\*
  \includegraphics[width=\subwidth]{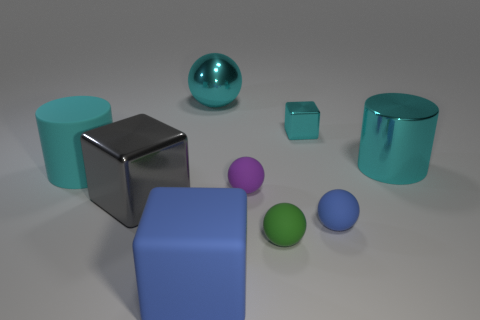} 
  \includegraphics[width=\subwidth]{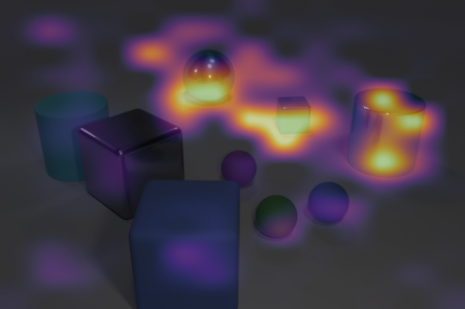} 
  \includegraphics[width=\subwidth]{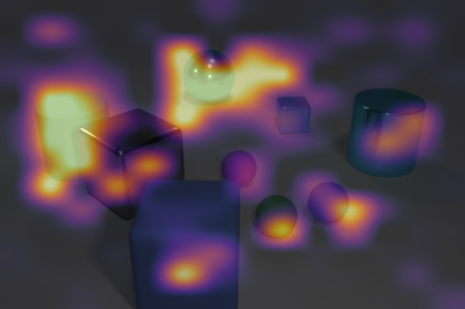} 
  \includegraphics[width=\subwidth]{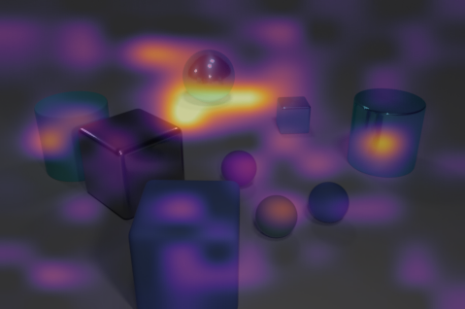} 
  \includegraphics[width=\subwidth]{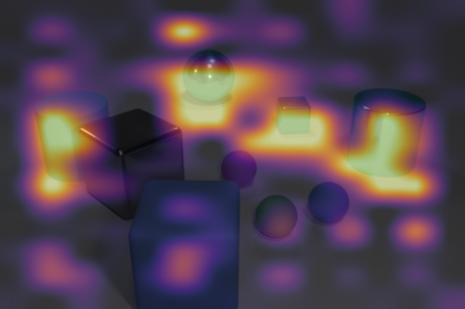} \\*
  \begin{minipage}{\subwidth}
  \small\centering \textbf{Q:} \textit{How many \textcolor{cyan}{cyan}\\things are}\dots \\~\\ \end{minipage}
  \begin{minipage}{\subwidth}
    \small\centering
    \dots\underline{\smash{right}}\textit{ of the }\underline{\smash{gray}}\textit{ cube?} \\~\\
    \textbf{A:} \textit{3}
  \end{minipage}
  \begin{minipage}{\subwidth}
    \small\centering
    \dots\underline{\smash{left}}\textit{ of the }\underline{small}\textit{ cube?} \\~\\
    \textbf{A:} \textit{2}
  \end{minipage}
  \begin{minipage}{\subwidth}
    \small\centering
    \dots\textit{right of the gray cube }\underline{and}\textit{ left of the small cube?} \\*
    \textbf{A:} \textit{1}
  \end{minipage}
  \begin{minipage}{\subwidth}
    \small\centering
    \dots\textit{right of the gray cube }\underline{or}\textit{ left of the small cube?} \\*
    \textbf{A:} \textit{4}
  \end{minipage}
  \vspace{2mm}
  \caption{
  Visualizations of the norm of the gradient of the sum of the predicted answer scores with respect to the final feature map. From left to right, each question adds a module to the program; the new module is \emph{\underline{underlined}} in the question. The visualizations illustrate which objects the model attends to when performing the reasoning steps for question answering. Images are from the validation set.}
  \label{fig:modules}
\end{figure*}

We evaluate our model on the recent CLEVR dataset~\cite{johnson2017clevr}. Standard VQA methods perform poorly on this dataset, showing that it is a challenging benchmark. All questions are equipped with ground-truth 
programs, allowing for experiments with varying amounts of supervision.

We first perform experiments using strong supervision in the form of ground-truth programs. We show that in this strongly supervised setting, the combination of \programmer and \interpreter works much better on CLEVR than alternative methods. Next, we show that this strong performance is maintained when a small number of ground-truth programs, which capture only a fraction of question diversity, is used for training. Finally, we evaluate the ability of our models to perform compositional generalization, as well as generalization to free-form questions posed by humans. Code reproducing the results of our experiments is available from \url{https://github.com/facebookresearch/clevr-iep}.

\subsection{Baselines}
Johnson \etal~\cite{johnson2017clevr} tested several VQA models on CLEVR.
We reproduce these models as baselines here.

\textbf{Q-type mode:} This baseline predicts the most frequent
answer for each of the question types in CLEVR.

\textbf{LSTM:}
Similar to \cite{antol2015vqa,malinowski15neurons},
questions are processed with learned word embeddings followed by a
word-level LSTM~\cite{hochreiter97}. The final LSTM hidden state is passed
to a multi-layer perceptron (MLP) that predicts a distribution over answers.
This method uses no image information, so it can only model
question-conditional biases.

\textbf{CNN+LSTM:}
Images and questions are encoded using convolutional network (CNN) features and final LSTM hidden states, respectively.
These features are concatenated and passed to a MLP that predicts an answer distribution.

\textbf{CNN+LSTM+SA~\cite{yang16stackedattention}:} Questions and images are encoded using a CNN and LSTM
as above, then combined using two rounds of soft spatial
attention; a linear transform of the attention output predicts the answer.

\textbf{CNN+LSTM+SA+MLP:} Replaces the linear transform with an MLP for better comparison with the
other methods.

The models that are most similar to ours are neural module networks~\cite{andreas2016learning,andreas2016neural}.
Unfortunately, neural module networks use a hand-engineered, off-the-shelf parser to produce programs, and this parser fails\footnote{See supplemental material for example parses of CLEVR questions.} on the complex questions in CLEVR~\cite{johnson2017clevr}. Therefore, we were unable to include module networks in our experiments.

\begin{figure}
  \centering
  \includegraphics[height=0.1185\textwidth]{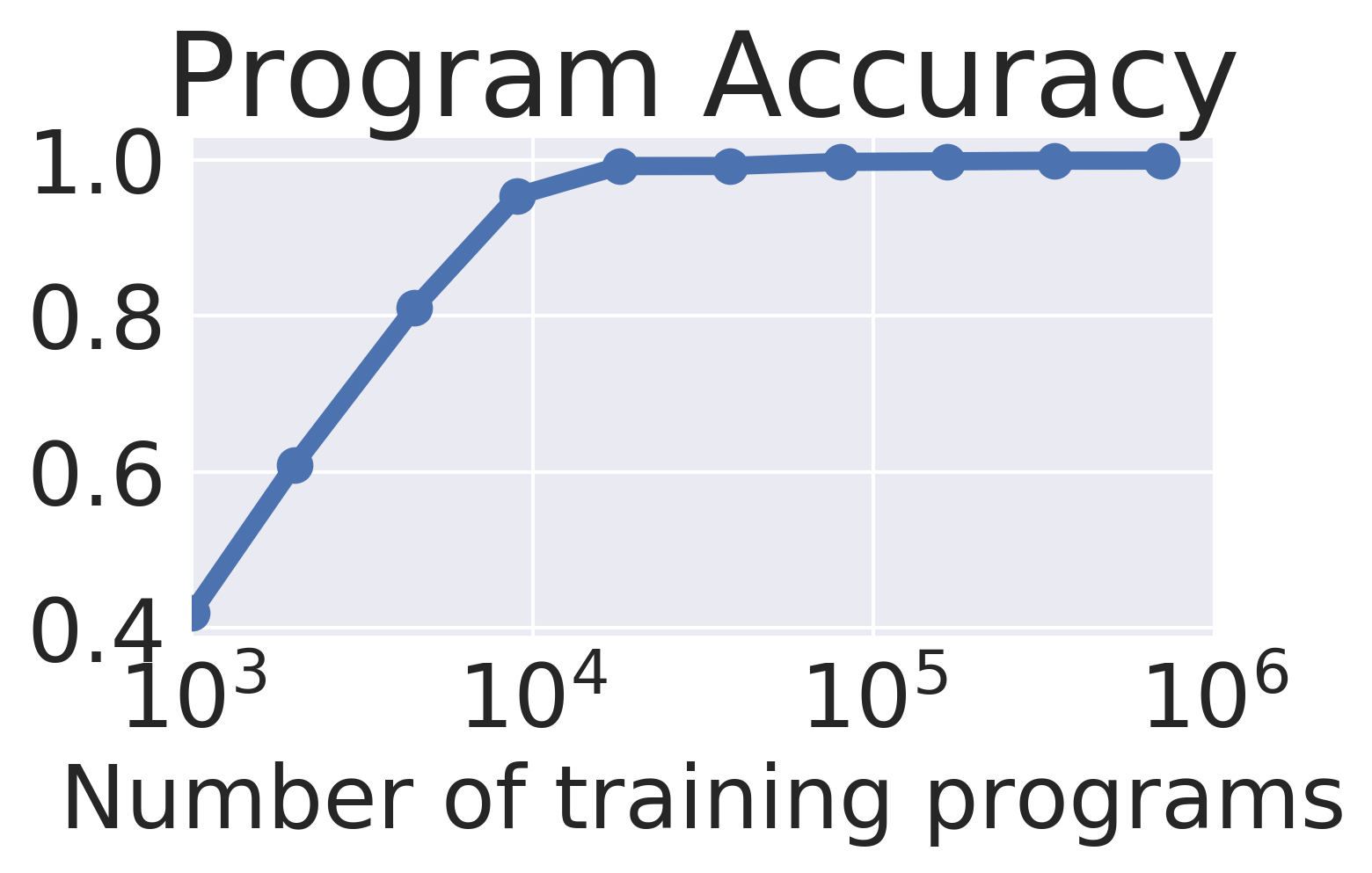}
  \includegraphics[height=0.1185\textwidth]{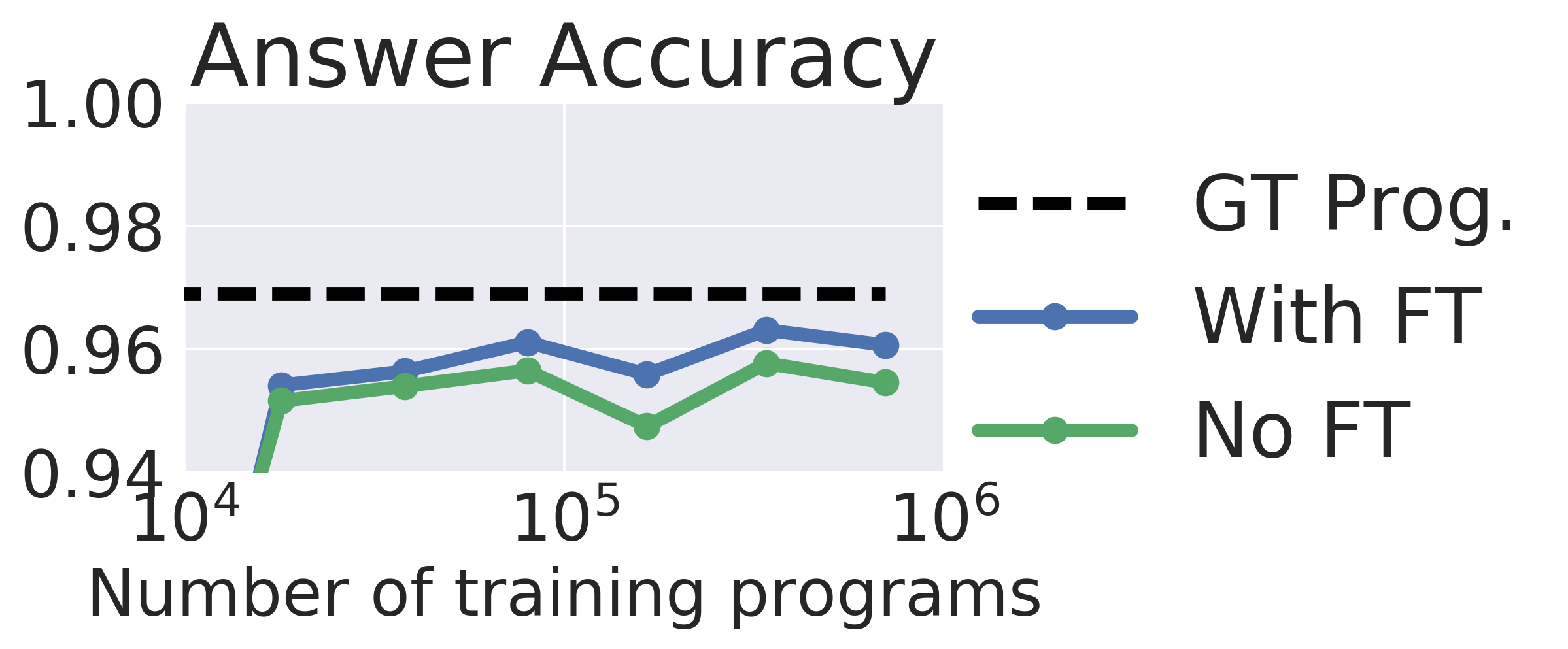}
  \caption{
    Accuracy of predicted programs (left) and answers (right) as we vary the number of
    ground-truth programs. Blue and green give accuracy before and after joint finetuning;
    the dashed line shows accuracy of our strongly-supervised model.
  }
 \label{fig:accvsdata}
\end{figure}

\begin{figure}
  \small
  \centering
  \begin{tabular}{r|cc|cc}
                     & \multicolumn{2}{c|}{Train A} & \multicolumn{2}{c}{Finetune B} \\
    Method           &    A    &   B   &    A  & B \\
    \hline\hline
    LSTM             &  55.2  & 50.9 & 51.5 & 54.9 \\
    CNN+LSTM         &  63.7  & 57.0 & 58.3 & 61.1 \\
    CNN+LSTM+SA+MLP    &  80.3  & 68.7 & 75.7 & 75.8 \\
    Ours (18K prog.) &  \textbf{96.6}  & \textbf{73.7} & \textbf{76.1} & \textbf{92.7}
  \end{tabular}
  \includegraphics[width=0.45\textwidth]{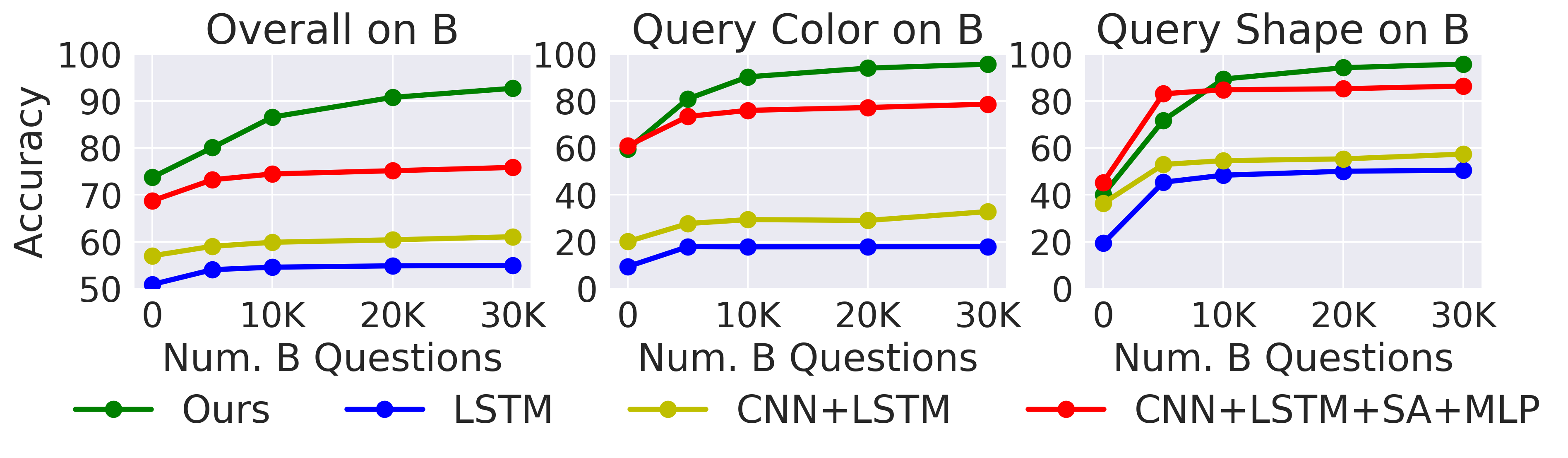}
  \vspace{1mm}
  \caption{
    Question answering accuracy on the CLEVR-CoGenT dataset (higher is better).
    \textbf{Top:} We train models on Condition A, then test them on both
    Condition A and Condition B. We then finetune these models on Condition B
    using 3K images and 30K questions, and again test on both Conditions. Our
    model uses 18K programs during training on Condition A, and does not use any
    programs during finetuning on Condition B.
    \textbf{Bottom:} We investigate the effects of using different amounts of
    data when finetuning on Condition B. We show overall accuracy as well as
    accuracy on color-query and shape-query questions.
  }
  \label{tab:cogent}
\end{figure}

\subsection{Strongly and semi-supervised learning}
\label{sec:strong-supervision}
We first experiment with a model trained using full supervision: we use the ground-truth programs for all questions in CLEVR to train both the \programmer and the \interpreter separately. The question answering accuracy of the resulting model on CLEVR is shown in Table~\ref{tab:fullclevr} (Ours-strong). The results show that using strong supervision, our model can achieve near-perfect accuracy on CLEVR (even outperforming Mechanical Turk workers).

In practical scenarios, ground-truth programs are not available for all questions. 
We use the semi-supervised training process described in Section~\ref{sec:training} to
determine how many ground-truth programs are needed to match fully supervised models.
First, the \programmer is trained in a supervised manner using a small number of questions
and ground-truth programs; next, the \interpreter is trained on \emph{all} CLEVR questions,
using predicted rather than ground-truth programs. Finally, both components are jointly
finetuned \emph{without} ground-truth programs. Table~\ref{tab:fullclevr} shows the accuracy
of semi-supervised models trained with 9K and 18K ground-truth programs (Ours-semi).

The results show that 18K ground-truth programs are sufficient to train a model that performs almost on par with a fully supervised model (that used all 700K programs for training). This strong performance is not due to the \programmer simply remembering all programs: the total number of unique programs in CLEVR is approximately 450K. This implies that after observing only a small fraction ($\leq\!4\%$) of all possible programs, the model is able to understand the underlying structure of CLEVR questions and use that understanding to generalize to new questions.

Figure~\ref{fig:accvsdata} analyzes how the accuracy of the predicted programs and the final answer vary with the number of ground-truth programs used.
We measure the accuracy of the \programmer by deserializing the function sequence produced by the \programmer, and marking it as correct if it matches the ground-truth program exactly.\footnote{Note that this may underestimate the true accuracy, since two different programs can be functionally equivalent.} 
Our results show that  with about 20K ground-truth programs, the \programmer achieves near perfect accuracy, and the final answer accuracy is almost as good as strongly-supervised training. 
Training the \interpreter using the predicted programs from the \programmer instead of ground-truth programs leads to a loss of about $3$ points in accuracy, but some of that loss is mitigated after joint finetuning.

\subsection{What do the modules learn?}
To obtain additional insight into what the modules in the \interpreter have learned, we visualized the parts of the image that are being used to answer different questions; see Figure~\ref{fig:modules}. Specifically, the figure displays the norm of the gradient of the sum of the predicted answer scores (softmax inputs) with respect to the final feature map. This visualization reveals several important aspects of our model.

First, it clearly attends to the correct objects even for complicated referring expressions involving spatial relationships, intersection and union of constraints, \emph{etc.}

Second, the examples show that changing a \emph{single} module (swapping
\emph{purple}/\emph{blue}, \emph{left}/\emph{right}, \emph{and}/\emph{or})
results in drastic changes in both the predicted answer and model attention,
demonstrating that the individual modules do in fact perform their intended functions.
Modules learn specialized functions such as localization and set operations without
explicit supervision of their outputs.

\begin{figure}
  \centering
  \begin{minipage}{0.22\textwidth}
    \footnotesize
    \centering
    \textbf{Ground-truth question}:\\*
    \textit{Is the number of matte blocks in front of the small yellow cylinder greater than the number of
    red rubber spheres to the left of the large red shiny cylinder?} \\*
    \textbf{Program length:} 20 \hspace{2mm}
    \textbf{A:} \textit{yes}~\cmark
  \end{minipage}
  \begin{minipage}{0.22\textwidth}
    \footnotesize
    \centering
    \textbf{Ground-truth question}:\\*
    \textit{How many objects are big rubber objects that are in front of the big gray thing or large rubber things that are in front of the large rubber sphere?} \\*
    \textbf{Program length:} 16 \hspace{2mm}
    \textbf{A:} \textit{1}~\cmark \hspace{1mm}
  \end{minipage} \\*
  \includegraphics[width=0.22\textwidth]{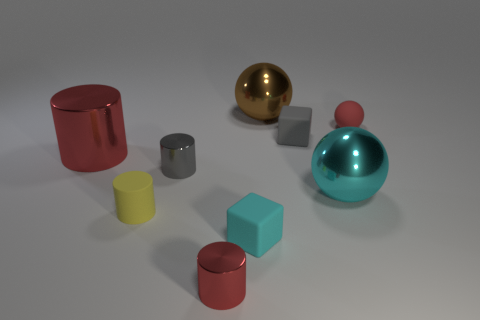}
  \includegraphics[width=0.22\textwidth]{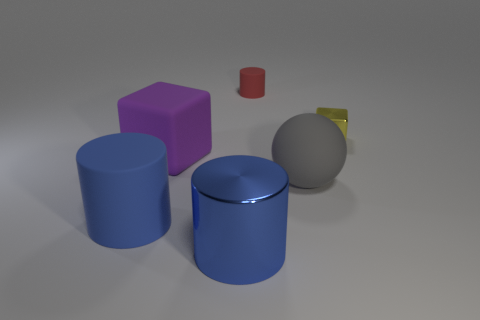}
  \begin{minipage}{0.22\textwidth}
    \footnotesize
    \centering
    \textbf{Predicted program} {\scriptsize (translated)}: \\*
    \textit{Is the number of matte blocks in front of the small yellow cylinder greater than the number of large red shiny cylinders?} \\*
    \textbf{Program length:} 15 \hspace{2mm}
    \textbf{A:} \it{no}~\xmark
  \end{minipage}
  \begin{minipage}{0.22\textwidth}
    \footnotesize
    \centering
    \textbf{Predicted program} {\scriptsize (translated)}: \\*
    \textit{How many objects are big \\* rubber objects in front of the \\* big gray thing or large \\* rubber spheres?} \\*
    \textbf{Program length:} 12 \hspace{2mm}
    \textbf{A:} \textit{2}~\xmark
  \end{minipage}
  \vspace{2mm}
  \caption{
    Examples of long questions where the program and answer were predicted incorrectly when
    the model was trained on short questions, but both program and answer were correctly predicted after
    the model was finetuned on long questions. Above each image we show the ground-truth question and its 
    program length; below, we show a manual English translation of the predicted program and answer before
    finetuning on long questions.
  }
  \label{fig:shortvslong}
\end{figure}

\begin{table}
  \small
  \centering
  \begin{tabular}{r|cc|cc}
    & \multicolumn{2}{c|}{Train Short} & \multicolumn{2}{c}{Finetune Both} \\
    Method & Short & Long & Short & Long \\
    \hline\hline
    LSTM & 46.4 & 48.6 & 46.5 & 49.9 \\
    CNN+LSTM & 54.0 & 52.8 & 54.3 & 54.2 \\
    CNN+LSTM+SA+MLP & 74.2 & \bf{64.3} & 74.2 & 67.8 \\
    Ours (25K prog.) & \bf{95.9} & 55.3 & \bf{95.6} & \bf{77.8} \\
  \end{tabular}
  \vspace{1mm}
  \caption{
    Question answering accuracy on short and long CLEVR questions. \textbf{Left columns}: Models trained only on short
    questions; our model uses 25K ground-truth short programs. \textbf{Right columns}: Models trained on both short and
    long questions. Our model is trained on short questions then finetuned on the entire dataset; no ground-truth programs
    are used during finetuning.
  }
  \label{tab:question-split}
\end{table}

\subsection{Generalizing to new attribute combinations}
Johnson \etal~\cite{johnson2017clevr} proposed the CLEVR-CoGenT dataset for investigating the ability of VQA models to perform compositional generalization. The dataset contains data in two different conditions: in Condition A, all cubes are
gray, blue, brown, or yellow and all cylinders are red, green,
purple, or cyan; in Condition B, cubes and cylinders swap color palettes. Johnson \etal~\cite{johnson2017clevr} found that VQA models trained on data from Condition A performed poorly on data from Condition B, suggesting the models are not well capable of generalizing to new conditions.

We performed experiments with our model on CLEVR-CoGenT: in Figure~\ref{tab:cogent}, we report accuracy of the semi-supervised variant of our model trained on data from Condition A and evaluated on data from Condition B.
Although the resulting model performs better than all baseline methods in Condition B, it still appears to suffer from the problems identified by \cite{johnson2017clevr}. A more detailed analysis of the results revealed that our model does not outperform the CNN+LSTM+SA baseline for questions about an object's shape or color. This is not surprising: if the model never sees red cubes, it has no incentive to learn that the attribute ``red'' refers to the color and not to the shape.

We also performed experiments in which we used a small amount of training data \emph{without ground-truth programs} from condition B for finetuning. 
We varied the amount of data from condition B that is available for finetuning. 
As shown in Figure~\ref{tab:cogent}, our model learns the new attribute combinations from only $\sim$10K questions ($\sim$1K images), and outperforms similarly trained baselines across the board.\footnote{Note that this finetuning hurts performance on condition $A$. Joint finetuning on both conditions will likely alleviate this issue.}
We believe that this is because the model's compositional nature allows it to quickly learn new semantics of attributes such as ``red" from little training data.

\subsection{Generalizing to new question types}
Our experiments in Section~\ref{sec:strong-supervision} showed that relatively few ground-truth programs are required
to train our model effectively. Due to the large number of unique programs in CLEVR, it is impossible to
capture all possible programs with a small set of ground-truth programs; however, due to the synthetic
nature of CLEVR questions, it is possible that a small number of programs could cover all possible
program structures. In real-world scenarios, models should be able to generalize to questions with
novel program structures without observing associated ground-truth programs.

To test this, we divide CLEVR questions into two categories based on their ground-truth programs:
\emph{short} and \emph{long}. CLEVR questions are divided into \emph{question families}, where all
questions in the same family share the same program structure. A question is \emph{short} if its
question family has a mean program length less than 16; otherwise it is \emph{long}.\footnote{Partitioning
at the family level rather than the question level allows for better separation of program structure
between short and long questions.}

We train the \programmer and \interpreter on short questions in a semi-supervised manner using 18K
ground-truth short programs, and test the resulting model on both short and long questions. This experiment
tests the ability of our model to generalize from short to long chains of reasoning. Results are shown in
Table~\ref{tab:question-split}.

The results show that when evaluated on long questions, our model trained on short questions underperforms the CNN+LSTM+SA model trained on the same set. Presumably, this result is due to the \programmer learning a bias towards short programs. Indeed, Figure~\ref{fig:shortvslong} shows that the \programmer produces programs that refer to the right objects but that are too short.

We can undo this short-program bias through joint finetuning of the \programmer and \interpreter on the
combined set of short and long questions, \emph{without ground-truth programs}. To pinpoint the problem of
short-program bias in the \programmer, we leave the \interpreter fixed during finetuning; it is only
used to compute REINFORCE rewards for the \programmer. After finetuning, our model substantially
outperforms baseline models that were trained on the entire dataset; see Table~\ref{tab:question-split}.

\begin{table}
  \small
  \centering
  \scalebox{0.95}{
  \begin{tabular}{r|cc}
           & Train & Train CLEVR, \\
    Method & CLEVR & finetune human \\
    \hline\hline
    LSTM & 27.5 & 36.5 \\
    CNN+LSTM & 37.7 & 43.2 \\
    CNN+LSTM+SA+MLP & 50.4 & 57.6 \\
    Ours (18K prog.) & \bf{54.0} & \bf{66.6} \\
  \end{tabular}
  }
  \vspace{1mm}
  \caption{
    Question answering accuracy on the CLEVR-Humans test set of four models after training on just the CLEVR dataset (\textbf{left}) and after finetuning on the CLEVR-Humans dataset (\textbf{right}).
  }\label{table:human_perf}
  \vspace{-3mm}
\end{table}

\begin{figure*}
  \begin{minipage}[t]{0.20\textwidth}
    \centering
    \footnotesize
    \includegraphics[width=0.95\textwidth]{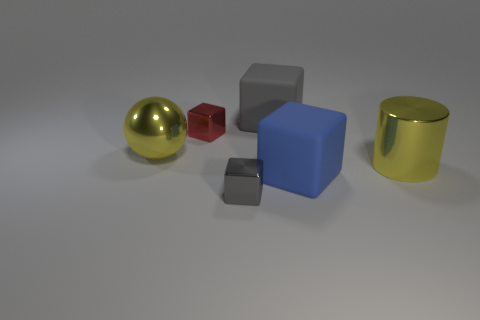}
    \textbf{Q:} \textit{Is there a blue }\underline{box}\textit{ \\* in the }\underline{items}\textit{?}
    \textbf{A:} \textit{yes} \\*[7.5mm]
    \textbf{Predicted Program:} \\*
    \textcolor{OliveGreen}{
      \bf
      \texttt{exist} \\*
      \texttt{filter\_shape[cube]} \\*
      \texttt{filter\_color[blue]} \\*
      \texttt{scene} \\*
    }
    \vspace{19.5mm}
    \textbf{Predicted Answer:} \\*
    \cmark~\textit{yes}
  \end{minipage}%
  \begin{minipage}[t]{0.20\textwidth}
    \centering
    \footnotesize
    \includegraphics[width=0.95\textwidth]{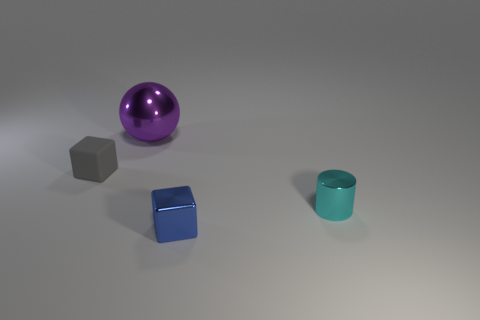}
    \textbf{Q:} \textit{What shape object \\* is }\underline{\smash{farthest}}\textit{ right?} \\*
    \textbf{A:} \textit{cylinder} \\*[4mm]
    \textbf{Predicted Program:} \\*
    \textcolor{BurntOrange}{
      \bf
      \texttt{query\_shape} \\*
      \texttt{unique} \\*
      \texttt{relate[right]} \\*
      \texttt{unique} \\*
      \texttt{filter\_shape[cylinder]} \\*
      \texttt{filter\_color[blue]} \\*
      \texttt{scene} \\*
    }
    \vspace{9.75mm}
    \textbf{Predicted Answer:} \\*
    \cmark~\textit{cylinder}
  \end{minipage}%
  \begin{minipage}[t]{0.20\textwidth}
    \centering
    \footnotesize
    \includegraphics[width=0.95\textwidth]{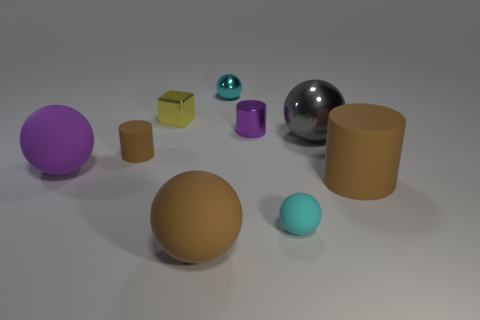}
    \textbf{Q:} \textit{Are }\underline{all}\textit{ the balls small?} \\*
    \textbf{A:} \textit{no} \\*[7.2mm]
    \textbf{Predicted Program:} \\*
    \textcolor{BurntOrange}{
      \bf
      \texttt{equal\_size} \\*
      \texttt{query\_size} \\*
      \texttt{unique} \\*
      \texttt{filter\_shape[sphere]} \\*
      \texttt{scene} \\*
      \texttt{query\_size} \\*
      \texttt{unique} \\*
      \texttt{filter\_shape[sphere]} \\*
      \texttt{filter\_size[small]} \\*
      \texttt{scene} \\*
    }
    \textbf{Predicted Answer:} \\*
    \cmark~\textit{no}
  \end{minipage}%
  \begin{minipage}[t]{0.20\textwidth}
    \centering
    \footnotesize
    \includegraphics[width=0.95\textwidth]{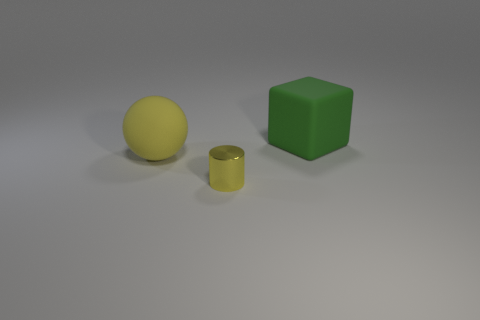}
    \textbf{Q:} \textit{Is the green block to the \\* right of the yellow sphere?} \\*
    \textbf{A:} \textit{yes} \\*[4mm]
    \textbf{Predicted Program:} \\*
    \textcolor{BurntOrange}{
      \bf
      \texttt{exist} \\*
      \texttt{filter\_shape[cube]} \\*
      \texttt{filter\_color[green]} \\*
      \texttt{relate[right]} \\*
      \texttt{unique} \\*
      \texttt{filter\_shape[sphere]} \\*
      \texttt{filter\_color[yellow]} \\*
      \texttt{scene} \\*
    }
    \vspace{6.5mm}
    \textbf{Predicted Answer:} \\*
    \cmark \textit{yes}
  \end{minipage}%
  \begin{minipage}[t]{0.2\textwidth}
    \centering
    \footnotesize
    \includegraphics[width=0.95\textwidth]{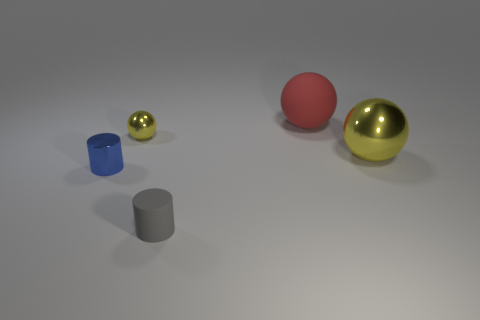}
    \textbf{Q:} 
      \underline{Two} \underline{items} \underline{share}\textit{ a color, a \\* material,
      and a shape; what \\* is the size of the }\underline{\smash{rightmost}}\textit{ \\* of }
      \underline{those} \underline{items}?
    \textbf{A:} \textit{large} \\*[0.75mm]
    \textbf{Predicted Program:} \\*
    \textcolor{BrickRed}{
      \bf
      \texttt{count} \\*
      \texttt{filter\_shape[cube]} \\*
      \texttt{same\_material} \\*
      \texttt{unique} \\*
      \texttt{filter\_shape[cylinder]} \\*
      \texttt{scene} \\*
    }
    \vspace{13mm}
    \textbf{Predicted Answer:} \\*
    \xmark~\textit{0}
  \end{minipage}
  \vspace{0.5mm}
  \caption{
    Examples of questions from the CLEVR-Humans dataset, along with predicted programs
    and answers from our model. Question words that do not appear in CLEVR questions are
    \underline{underlined}. Some predicted programs exactly match the semantics of the question
    (\textcolor{OliveGreen}{green}); some programs closely match the question semantics
    (\textcolor{BurntOrange}{yellow}), and some programs appear unrelated to the question
    (\textcolor{BrickRed}{red}).
  }
  \label{fig:human-examples}
\end{figure*}

\subsection{Generalizing to human-posed questions}
The fact that questions in the CLEVR benchmark were generated algorithmically may favor some approaches over others.
In particular, natural language tends to be more ambiguous than algorithmically generated questions. We performed an experiment to assess the extent to which models trained on CLEVR can be finetuned to answer human questions. To this end, we collected
 a new dataset of natural-language questions and answers for CLEVR images.

\paragraph{The CLEVR-Humans Dataset.} 
Inspired by VQA~\cite{antol2015vqa}, workers on Amazon Mechanical Turk were asked to write questions about CLEVR images
that would be \emph{hard for a smart robot to answer};
workers were primed with questions from CLEVR and restricted to answers in CLEVR.
We filtered questions by asking three workers to answer each question, and removed questions that
a majority of workers could not correctly answer. We collected one question per image; after filtering, we obtained
17,817 training, 7,202 validation, and 7,145 test questions on CLEVR images.
The data is available from the first author's website.

The human questions are more challenging than synthetic CLEVR questions because
they exhibit more linguistic variety. Unlike existing VQA datasets, however,
the CLEVR-Humans questions do not require common-sense knowledge: they focus entirely on visual reasoning abilities, which makes them a good testbed for evaluating reasoning.

Figure~\ref{fig:human-examples} shows some example human questions.
Some questions are rewordings of synthetic CLEVR questions; others are answerable using
the same basic functions as CLEVR but potentially with altered semantics for those
skills. For example, people use spatial relationships ``left'', ``right'', \emph{etc.}
differently than their meanings in CLEVR questions. Finally, some questions require
skills not needed for answering synthetic questions.

\paragraph{Results.}
We train our model on CLEVR, and then finetune \emph{only the \programmer} on the CLEVR-Humans training set to adapt it to the additional linguistic variety; we do not adapt the \interpreter due to the limited quantity of data.
No ground-truth programs are available during finetuning.
The embeddings in the sequence-to-sequence model of question words that do not appear in CLEVR synthetic questions are initialized randomly before finetuning.

During finetuning, our model learns to reuse the reasoning skills it has already mastered in order to answer
the linguistically more diverse natural-language questions. As shown in Figure~\ref{fig:human-examples}, it
learns to map novel words (\emph{``box''}) to known modules. When human questions are not expressible using CLEVR
functions, our model still learns to produce reasonable programs closely approximating the question's intent.
Our model often fails on questions that cannot be reasonably approximated using our model's module inventory,
such as the rightmost example in Figure~\ref{fig:human-examples}.
Quantitatively, the results in Table~\ref{table:human_perf} show that our model outperforms all baselines on the CLEVR-Humans test set both with and
without finetuning.

\section{Discussion and Future Work}
\label{sec:discussion}
Our results show that our model is able to generalize to novel scenes and questions and can even
infer programs for free-form human questions using its learned modules. Whilst these results are encouraging, there still are many questions that cannot be reasonably approximated
using our fixed set of modules. For example, the question \emph{``What color is the object with a
unique shape?''} requires a model to identify unique shapes, for which no module is currently available. Adding new modules to our model
is straightforward due to our generic module design, but automatically identifying and learning new
modules without program supervision is still an open problem. One path forward is to design a Turing-complete set of modules; this would allow for all
programs to be expressed without learning new modules. For example, by adding ternary
operators~({\small\tt if/then/else}) and loops~({\small\tt for/do}), the question
\emph{``What color is the object with a unique shape?''} can be answered by \emph{looping}
over all shapes, counting the objects with that shape, and returning it \emph{if} the count is one. These control-flow operators could be incorporated into our
framework: for example, a loop could apply the same module to an input set and aggregate the results.
We emphasize that learning such programs with limited supervision is an open
research challenge, which we leave to future work.

\section{Conclusion}
This paper fits into a long line of work on incorporating symbolic representations into (neural) machine learning models \cite{balog2017deepcoder,cai2017recursion,liang2016neural,reed2016neural}. We have shown that explicit program representations can make it easier to compose programs to answer novel questions about images. Our generic program representation, learnable \programmer and universal design for modules makes our model much more flexible than neural module networks~\cite{andreas2016learning,andreas2016neural} and thus more easily extensible to new problems and domains.

\clearpage
\appendix
\begin{figure}[ht!]
  \centering
  \Large\textbf{Supplementary Material}
\end{figure}
\section{Implementation Details}

We will release code to reproduce our experiments. We also detail some key
implementation details here.

\subsection{Program Generator}
In all experiments our program generator is an LSTM sequence-to-sequence model~\cite{sutskever2014sequence}. It comprises two learned recurrent neural networks: the \emph{encoder} receives the natural-language question as a sequence of words, and summarizes the question as a fixed-length vector; the \emph{decoder} receives this fixed-length vector as input and produces the predicted program as a sequence of functions. The encoder and decoder do not share weights.

The encoder converts the discrete words of the input question to vectors of dimension 300 using a learned word embedding layer; the resulting sequence of vectors is then processed with a two-layer LSTM using 256 hidden units per layer. The hidden state of the second LSTM layer at the final timestep is used as the input to the decoder network.

At each timestep the decoder network receives both the function from the previous timestep (or a special \texttt{<START>} token at the first timestep) and the output from the encoder network. The function is converted to a 300-dimensional vector with a learned embedding layer and concatenated with the decoder output; the resulting sequence of vectors is processed by a two-layer LSTM with 256 hidden units per layer. At each timestep the hidden state of the second LSTM layer is used to compute a distribution over all possible functions using a linear projection.

During supervised training of the program generator, we use
Adam~\cite{kingma2014adam} with a learning rate of $5\times10^{-4}$ and a batch
size of 64; we train for a maximum of 32,000 iterations, employing early stopping
based on validation set accuracy.

\subsection{Execution Engine}
The execution engine uses a Neural Module Network~\cite{andreas2016neural} to
compile a custom neural network architecture based on the predicted program from 
the program generator. The input image is first resized to $224\times224$ pixels,
then passed through a convolutional network to extract image features; the
architecture of this network is shown in Table~\ref{table:cnn-arch}.

\begin{table}
  \centering
  \begin{tabular}{|c|c|}
    \hline
    Layer & Output size \\
    \hline
    Input image & $3\times224\times224$ \\
    ResNet-101~\cite{he2016deep} conv4\_6 & $1024\times14\times14$ \\
    Conv($3\times3$, $1024\to128$) & $128\times14\times14$ \\
    ReLU & $128\times14\times14$ \\
    Conv($3\times3$, $128\to128$) & $128\times14\times14$ \\
    ReLU & $128\times14\times14$ \\
    \hline
  \end{tabular}
  \vspace{1mm}
  \caption{
    Network architecture for the convolutional network used in our execution
    engine. The ResNet-101 model is pretrained on
    ImageNet~\cite{russakovsky2015imagenet} and remains fixed
    while the execution engine is trained. The output from this network is
    passed to modules representing \texttt{Scene} nodes in the program.
  }
  \label{table:cnn-arch}
  \vspace{-4mm}
\end{table}

The predicted program takes the form of a syntax tree; the leaves of the tree
are \texttt{Scene} functions which receive visual input from the convolutional
network. For ground-truth programs, the root of the tree is a function
corresponding to one of the question types from the CLEVR
dataset~\cite{johnson2017clevr}, such as \texttt{count} or
\texttt{query\_shape}. For predicted programs the root of the program tree could
in principle be any function, but in practice we find that trained models tend
only to predict as roots those function types that appear as roots of
ground-truth programs.

Each function in the predicted program is associated with a \emph{module}
which receives either one or two inputs; this association gives rise to a
custom neural network architecture corresponding to each program. Previous
implementations of Neural Module
networks~\cite{andreas2016learning,andreas2016neural} used different
architectures for each module type, customizing the module architecture to the
function the module was to perform. In contrast we use a generic design for our
modules: each module is a small residual block~\cite{he2016deep}; the exact
architectures used for our unary and binary modules are shown in
Tables~\ref{table:unary-module-arch} and \ref{table:binary-module-arch}
respectively.

In initial experiments we used Batch Normalization~\cite{ioffe2015batch}
after each convolution in the modules, but we found that this prevented the
model from converging. Since each image in a minibatch may have a different
program, our implementation of the execution engine iterates over each program
in the minibatch one by one; as a result each module is only run with a batch
size of one during training, leading to poor convergence when modules contain
Batch Normalization.

The output from the final module is passed to a classifier which predicts
a distribution over answers; the exact architecture of the classifier is shown
in Table~\ref{table:classifier-arch}.

When training the execution engine alone (using either ground-truth programs
or predicted programs from a fixed program generator), we train using
Adam~\cite{kingma2014adam} with a learning rate of $1\times10^{-4}$ and a batch
size of 64; we train for a maximum of 200,000 iterations and employ early
stopping based on validation set accuracy.

\begin{table}
  \centering
  \begin{tabular}{|c|c|c|}
    \hline
    Index & Layer & Output size \\
    \hline
    (1) & Previous module output & $128\times14\times14$ \\
    (2) & Conv($3\times3$, $128\to128$) & $128\times14\times14$ \\
    (3) & ReLU & $128\times14\times14$ \\
    (4) & Conv($3\times3$, $128\to128$) & $128\times14\times14$ \\
    (5) & Residual: Add (1) and (4) & $128\times14\times14$ \\
    (6) & ReLU & $128\times14\times14$ \\
    \hline
  \end{tabular}
  \vspace{1mm}
  \caption{
    Architecture for unary modules used in the execution engine.
    These modules receive the output
    from one other module, except for the special \texttt{Scene} module
    which instead receives input from the convolutional network
    (Table~\ref{table:cnn-arch}).
  }
  \label{table:unary-module-arch}
\end{table}

\begin{table}
  \centering
  \begin{tabular}{|c|c|c|}
    \hline
    Index & Layer & Output size \\
    \hline
    (1) & Previous module output & $128\times14\times14$ \\
    (2) & Previous module output & $128\times14\times14$ \\
    (3) & Concatenate (1) and (2) & $256\times14\times14$ \\
    (4) & Conv($1\times1$, $256\to128$) & $128\times14\times14$ \\
    (5) & ReLU & $128\times14\times14$ \\
    (6) & Conv($3\times3$, $128\to128$) & $128\times14\times14$ \\
    (7) & ReLU & $128\times14\times14$ \\
    (8) & Conv($3\times3$, $128\to128$) & $128\times14\times14$ \\
    (9) & Residual: Add (5) and (8) & $128\times14\times14$ \\
    (10) & ReLU & $128\times14\times14$ \\
    \hline
  \end{tabular}
  \vspace{1mm}
  \caption{
    Architecture for binary modules in the execution engine.
    These modules receive the output from
    two other modules. The binary modules in our system are
    \texttt{intersect}, \texttt{union}, \texttt{equal\_size},
    \texttt{equal\_color}, \texttt{equal\_material}, \texttt{equal\_shape},
    \texttt{equal\_integer}, \texttt{less\_than}, and \texttt{greater\_than}.
  }
  \label{table:binary-module-arch}
\end{table}

\begin{table}
  \centering
  \begin{tabular}{|c|c|}
    \hline
    Layer & Output size \\
    \hline
    Final module output & $128\times14\times14$ \\
    Conv($1\times1$, $128\to512$) & $512\times14\times14$ \\
    ReLU & $512\times14\times14$ \\
    MaxPool($2\times2$, stride 2) & $512\times7\times7$ \\
    FullyConnected($512\cdot7\cdot7\to1024$) & $1024$ \\
    ReLU & $1024$ \\
    FullyConnected($1024\to|\mathcal{A}|$) & $|\mathcal{A}|$ \\
    \hline
  \end{tabular}
  \vspace{1mm}
  \caption{
    Network architecture for the classifier used in our execution engine.
    The classifier receives the output from the final module and predicts
    a distribution over answers $\mathcal{A}$.
  }
  \label{table:classifier-arch}
\end{table}

\subsection{Joint Training}
When jointly training the program generator and execution engine, we train
using Adam with a learning rate of $5\times10^{-5}$ and a batch size of 64;
we train for a maximum of 100,000 iterations, again employing early stopping
based on validation set accuracy.

We use a moving average baseline to reduce the variance of gradients estimated
using REINFORCE; in particular our baseline is an exponentially decaying moving
average of past rewards, with a decay factor of $0.99$.

\subsection{Baselines}
We reimplement the baselines used in~\cite{johnson2017clevr}:

\textbf{LSTM.} Our LSTM baseline receives the input question as a
sequence of words, converts the words to 300-dimensional vectors using a
learned word embedding layer, and processes the resulting sequence with a
two-layer LSTM with 512 hidden units per layer. The LSTM hidden state from
the second layer at the final timestep is passed to an MLP with two hidden
layers of 1024 units each, with ReLU nonlinearities after each layer.
\\*

\textbf{CNN+LSTM.} Like the LSTM baseline, the CNN+LSTM model encodes the
question using learned 300-dimensional word embeddings followed by a two-layer
LSTM with 512 hidden units per layer. The image is encoded using the same
CNN architecture as the execution engine, shown in Table~\ref{table:cnn-arch}.
The encoded question and (flattened) image features are concatenated and passed
to a two-layer MLP with two hidden layers of 1024 units each, with ReLU
nonlinearities after each layer.

\begin{table*}[ht!]
  \centering
  \begin{tabular}{|r|c|}
    \hline
    Question: & \begin{minipage}{0.7\textwidth}\centering\vspace{1mm}
      \emph{The brown object that is the same shape as the green \\*
            shiny thing is what size?}
    \vspace{1mm}\end{minipage} \\
    Fragments: & \texttt{(\_what \_thing)} \\
    \hline
    Question: & \emph{What material is the big purple cylinder?} \\
    Fragments: & \texttt{(material purple);(material big);(material (and purple big))} \\
    \hline
    Question: & \begin{minipage}{0.7\textwidth}\centering\vspace{1mm}
      \emph{How big is the cylinder that is in front of the green \\* 
            metal object left of the tiny shiny thing that is in \\* 
            front of the big red metal ball?}
    \vspace{1mm}\end{minipage} \\
    Fragments: & \texttt{(\_what \_thing)} \\
    \hline
    Question: & \begin{minipage}{0.7\textwidth}\centering\vspace{1mm}
      \emph{Are there any metallic cubes that are on the right side \\*
            of the brown shiny thing that is behind the small metallic  \\* 
            sphere to the right of the big cyan matte thing?}
    \vspace{1mm}\end{minipage} \\
    Fragments: & \texttt{(is brown);(is cubes);(is (and brown cubes))} \\
    \hline
    Question: & \begin{minipage}{0.7\textwidth}\centering\vspace{1mm}
      \emph{Is the number of cyan things in front of the purple matte \\* 
            cube greater than the number of metal cylinders left of the \\* 
            small metal sphere?}
    \vspace{1mm}\end{minipage} \\
    Fragments: & \texttt{(is cylinder);(is cube);(is (and cylinder cube))} \\
    \hline
    Question: & \emph{Are there more small blue spheres than tiny green things?} \\
    Fragments: & \texttt{(is blue);(is sphere);(is (and blue sphere))} \\
    \hline
    Question: & \emph{Are there more big green things than large purple shiny cubes?} \\
    Fragments: & \texttt{(is cube);(is purple);(is (and cube purple))} \\
    \hline
    Question: & \begin{minipage}{0.7\textwidth}\centering\vspace{1mm}
      \emph{What number of things are large yellow metallic balls or \\*
            metallic things that are in front of the gray metallic sphere?}
    \vspace{1mm}\end{minipage} \\
    Fragments: & \texttt{(number gray);(number ball);(number (and gray ball))} \\
    \hline
    Question: & \emph{The tiny cube has what color?} \\
    Fragments: & \texttt{(\_what \_thing)} \\
    \hline
    Question: & \begin{minipage}{0.7\textwidth}\centering\vspace{1mm}
      \emph{There is a small matte cylinder; is it the same color as the \\* 
            tiny shiny cube that is behind the large red metallic ball?}
    \vspace{1mm}\end{minipage} \\
    Fragments: & \texttt{(\_what \_thing)} \\
    \hline
  \end{tabular}
  \vspace{1mm}
  \caption{
    Examples of random questions from the CLEVR training set, parsed using the
    code by Andreas \etal~\cite{andreas2016learning} for parsing questions from
    the VQA dataset~\cite{antol2015vqa}. Each parse gives a set of \emph{layout fragments}
    separated by semicolons; in \cite{andreas2016learning} these fragments are combined
    to produce \emph{candidate layouts} for the module network. When the parser fails,
    it produces the default fallback fragment \texttt{(\_what \_thing)}.
  }
  \label{table:nmn2-parses}
\end{table*}

\textbf{CNN+LSTM+SA.} The question and image are encoded in exactly the same
manner as the CNN+LSTM baseline. However rather than concatenating these
representations, they are fed to two consecutive Stacked Attention
layers~\cite{yang16stackedattention} with a hidden dimension of 512 units;
this results in a 512-dimensional vector which is fed to a linear layer to
predict answer scores.

This matches the CNN+LSTM+SA model as originally
described by Yang \etal~\cite{yang16stackedattention}; this also matches
the CNN+LSTM+SA model used in \cite{johnson2017clevr}.

\textbf{CNN+LSTM+SA+MLP.} Identical to CNN+LSTM+ SA; however the output of the
final stacked attention module is fed to a two-layer MLP with
two hidden layers of 1024 units each, with ReLU nonlinearities after each layer.

Since all other other models (LSTM, CNN+LSTM, and ours) terminate in an MLP
to predict the final answer distribution, the CNN+LSTM+SA+MLP gives a more
fair comparison with the other methods.

Surprisingly, the minor architectural change of replacing the linear transform
with an MLP significantly improves performance on the CLEVR dataset: CNN+LSTM+SA
achieves an overall accuracy of 69.8, while CNN+LSTM+SA+MLP achieves 73.2. Much
of this gain comes from improved performance on comparison questions; for example
on shape comparison questions CNN+LSTM+SA achieves an accuracy of 50.9 and
CNN+LSTM+SA+MLP achieves 69.7.



\textbf{Training.} All baselines are trained using Adam with a learning rate of
$5\times10^{-4}$ with a batch size of 64 for a maximum of 360,000 iterations,
employing early stopping based on validation set accuracy.

\section{Neural Module Network parses}
The closest method to our own is that of Andreas \etal~\cite{andreas2016learning}.
Their dynamic neural module networks first perform a dependency parse of the
sentence; heuristics are then used to generate a set of \emph{layout fragments}
from the dependency parse. These fragments are heuristically combined, giving 
a set of \emph{candidate layouts}; the final network layout is selected from these
candidates through a learned reranking step.

Unfortunately we found that the parser used in~\cite{andreas2016learning} for VQA
questions did not perform well on the longer questions in CLEVR. In
Table~\ref{table:nmn2-parses} we show random questions from the CLEVR training set
together with the layout fragments computed using the parser from \cite{andreas2016learning}.
For many questions the parser fails, falling back to the fragment \texttt{(\_what \_thing)};
when this happens then the resulting module network will not respect the structure of the
question at all. For questions where the parser does not fall back to the default layout,
the resulting layout fragments often fail to capture key elements from the question; for
example, after parsing the question \emph{What material is the big purple cylinder?}, none of
the resulting fragments mention the \emph{cylinder}.

\clearpage

\paragraph{Acknowledgements.}
We thank Ranjay Krishna, Yuke Zhu, Kevin Chen, and Dhruv Batra for helpful
comments and discussion. J. Johnson is partially supported by an ONR MURI
grant.

{\small
\bibliographystyle{ieee}
\bibliography{iccv17}
}

\end{document}